\documentclass[conference]{IEEEtran}
\IEEEoverridecommandlockouts

\usepackage{amsmath,amssymb,amsfonts}
\usepackage{algorithmic}
\usepackage{graphicx}
\usepackage{textcomp}
\usepackage{xcolor}
\usepackage{bbm}
\usepackage{enumitem}
\usepackage{tabularx, booktabs}
\usepackage[style=ieee, maxcitenames=2, mincitenames=1, natbib=true, dashed=false]{biblatex}
\usepackage{soul,color}
\usepackage{siunitx}
\usepackage[capitalise]{cleveref}
\usepackage{dsfont}

\newcommand{\argmax}{\operatornamewithlimits{arg\,max}}

\begin{document}

\title{
    \LARGE \bf Robust Driving Policy Learning with \\ Guided Meta Reinforcement Learning
    
    \author{Kanghoon Lee$^{1,*}$, Jiachen Li$^{2,*}$, David Isele$^3$, Jinkyoo Park$^1$, Kikuo Fujimura$^3$, Mykel J. Kochenderfer$^{2}$}
    
    \thanks{$*$Equal Contribution}
    \thanks{$^1$K. Lee and J. Park are with the Systems Intelligence Laboratory (SILAB), Korea Advanced Institute of Science and Technology, South Korea. {\tt\small $\lbrace$leehoon, jinkyoo.park$\rbrace$@kaist.ac.kr}}
    \thanks{$^2$J. Li and M. J. Kochenderfer are with the Stanford Intelligent Systems Laboratory (SISL), Stanford University, CA, USA. {\tt\small $\lbrace$jiachen\_li, mykel$\rbrace$@stanford.edu}}
    \thanks{$^3$D. Isele and K. Fujimura are with the Honda Research Institute USA, CA, USA. {\tt\small $\lbrace$disele, kfujimura$\rbrace$@honda-ri.com}}
}
\maketitle

\begin{abstract}
Although deep reinforcement learning (DRL) has shown promising results for autonomous navigation in interactive traffic scenarios, existing work typically adopts a fixed behavior policy to control social vehicles in the training environment. This may cause the learned driving policy to overfit the environment, making it difficult to interact well with vehicles with different, unseen behaviors.
In this work, we introduce an efficient method to train diverse driving policies for social vehicles as a single meta-policy. 
By randomizing the interaction-based reward functions of social vehicles, we can generate diverse objectives and efficiently train the meta-policy through guiding policies that achieve specific objectives. 
We further propose a training strategy to enhance the robustness of the ego vehicle's driving policy using the environment where social vehicles are controlled by the learned meta-policy. 
Our method successfully learns an ego driving policy that generalizes well to unseen situations with out-of-distribution (OOD) social agents' behaviors in a challenging uncontrolled T-intersection scenario.  

\end{abstract}

\section{Introduction}
    \label{sec:intro}

Deep reinforcement learning has been a powerful tool for solving sequential decision making problems and has been applied in various domains, such as game playing \cite{lample2017playing}, robotics \cite{gu2017deep}, and autonomous driving \cite{ma2020latent, isele2018navigating}. 
DRL is characterized by its ability to handle high dimensional inputs due to the powerful representational capabilities of deep neural networks \cite{codevilla2018end, sadigh2016planning}, making it a strong tool for decision making in complex environments with intensive multi-agent interactions. 
However, DRL requires a large amount of data obtained by interacting with the environments to achieve satisfactory results. In addition, the learned policy is prone to performing poorly in out-of-distribution scenarios.

An open challenge in DRL for autonomous driving is to improve the robustness of the learned driving policies of autonomous vehicles (i.e., ego agents) to the variations in the driving policies of human-driven vehicles (i.e., social agents). 
In real-world settings, autonomous vehicles may be exposed to driving behaviors that are not necessarily similar to those they have seen during training. 
Existing methods for training DRL policies involve social agents whose policies are well-defined by the simulator. This is not ideal because simulated policies are fixed and not diverse, implying that the learned ego agent's policy tends to overfit the simulated behaviors of social agents. The challenge lies in finding a systematic way to generate new and diverse policies for social agents that could represent human-like behaviors to enhance the robustness of ego policies. 
We propose to train social agents' policies near the ego agent in the DRL training process with their own reward functions that involve a term that encourages cooperative or adversarial behaviors to the ego agent. 
The resulting social agents' policies are intended to imitate human drivers' behaviors while also challenging the ego agent. 

Another crucial component in DRL for autonomous driving is to accurately capture the internal preferences of other interactive agents, which could encapsulate their intents or driving traits. 
This is especially effective when the ego agent needs to negotiate the right of way with surrounding agents.
Capturing this internal preference is helpful for allowing the ego agent to make the most informed decisions. 
Prior work uses binary ground truth labels such as conservative or aggressive to describe the agent behaviors \cite{ma2020latent} and evaluates the proposed method in a driving simulator.
However, this approach is difficult to be deployed in a real-world setting since it is difficult to obtain the ground truth labels in practice. We use a variational auto-encoder (VAE) with a recurrent neural network (RNN) to encode behavior patterns in an unsupervised manner, which is similar to \cite{liu2022learning}. However, our VAE model is trained on a larger dataset that includes RL-based social vehicles.

\begin{figure}[!tbp]
    \centering
    \includegraphics[width=\columnwidth]{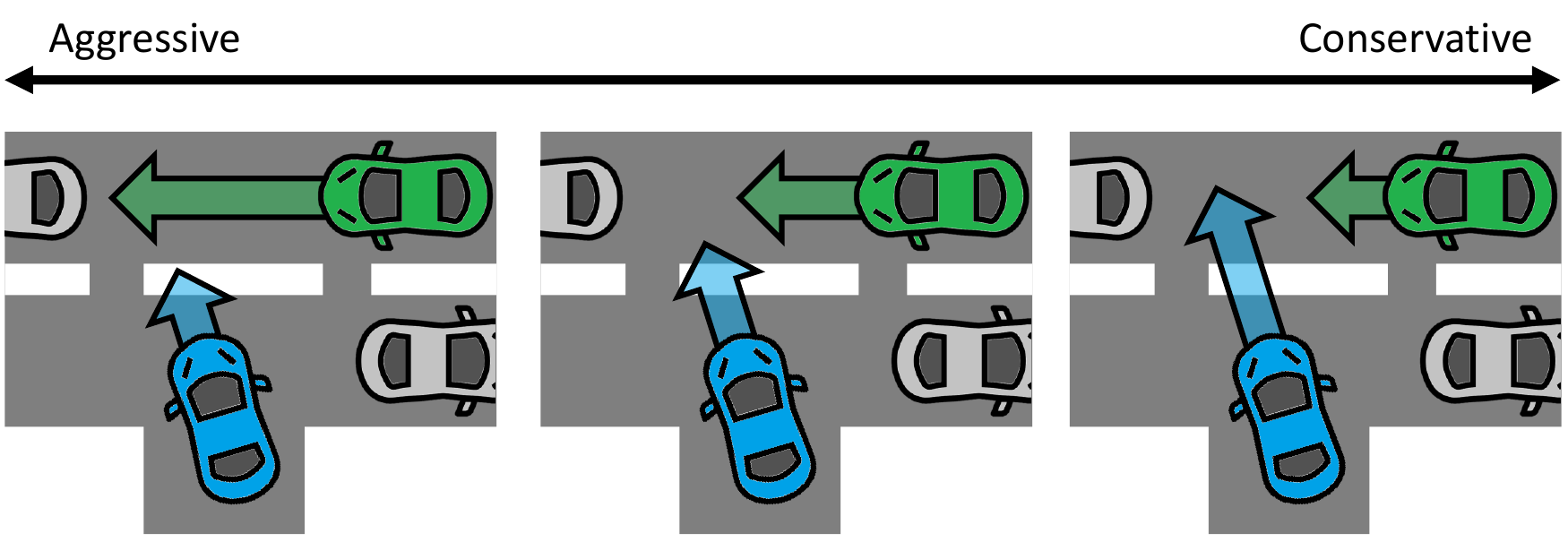}
    \vspace{-0.5cm}
    \caption{Social agents (i.e., green vehicles) exhibit different behaviors based on their preferences or internal characteristics in an uncontrolled T-intersection scenario, while the left-turn ego agent (i.e., blue vehicle) needs to react appropriately to their behaviors.}
    \vspace{-0.6cm}
    \label{fig:intro}
\end{figure}

The main contributions of the paper are as follows:
\begin{itemize}
    \item We propose a guided meta reinforcement learning method to generate diverse social behaviors, which includes training RL policies with different objectives as guiding policies and using them to improve the diversity of behaviors generated by the meta-RL policy. 
    \item We further propose to improve the robustness of the learned ego policy in a multi-agent reinforcement learning (MARL) framework by training the ego agent in environments with diverse social agents who have individual reward functions with a cooperative or adversarial component to encourage different levels of aggressiveness.
    \item We validate that the learned meta-policy generates behaviors associated with various objectives, and the learned ego driving policy achieves more robust performance in OOD situations than baseline methods in a challenging T-intersection scenario.
\end{itemize}

\section{Related Work}\label{sec:relatedwork}

\subsection{RL-Based Autonomous Driving}
Extensive research has been conducted on the intelligent navigation of autonomous vehicles through uncontrolled intersections. 
Existing approaches model the problem as a partially observable Markov decision process (POMDP) to account for the limited observability of the autonomous agent \cite{bouton2017belief, bretchel2014pomdp, wei2021autonomous}.  
For instance, \citet{bai2015intention} employ an unparameterized belief tracker, while \citet{song2016intention} use a hidden Markov model to capture driver movement intentions. 
However, the best-known algorithms for solving POMDPs are computationally expensive. 
Deep reinforcement learning methods have been effective in autonomous driving. 
\citet{isele2018navigating} apply deep Q-learning to navigate intersections with occluded vehicles. \citet{ma2020latent} employ graph neural networks and latent state inference to tackle the challenge of navigating through a T-intersection scenario. \citet{liu2022learning} extract an appropriate latent state from trajectory history without relying on the true latent state for the same scenario. 
Different from existing works, we specifically focus on the situations where the behavior policy of surrounding vehicles deviates from the training situations.

\subsection{Diverse Policy Learning}

Various studies have demonstrated the benefits of learning a collection of diverse policies across different aspects. The challenge of efficient exploration of RL problems is addressed by employing diverse different policies \cite{conti2017diversity, hong2018diversity, doan2019diversity, holder2020diversity}. 
\citet{mysore2022multicritic} introduce the concept of updating a single actor using multiple critics, each focusing on different objectives, to enable the incorporation of diverse behaviors. 
\citet{tang2021diverse, strouse2021collab} use a diverse set of training partners as an environment to train agents in the context of MARL, enabling them to collaborate effectively with unknown agents. 

Several techniques have been explored to generate a diverse policy set.
The first approach is through reward randomization, where the weight of the reward component is randomized, leading to each policy being trained using a distinct reward function \cite{tang2021diverse, mysore2022multicritic, chen2020lane}.
The second line of work uses distance metrics, such as the Kullback-Leibler (KL) divergence or mean squared error (MSE) loss, to quantify the dissimilarity between policy distributions, thereby distinguishing their behaviors \cite{lupu2021trajectory, doan2019diversity, hong2018diversity}.
The third line of work maximizes the diversity of the entire policy set by using a metric that quantifies the diversity of the set \cite{balduzzi2019open, holder2020diversity}.
While these methods are effective in generating a diverse policy set, they come with the drawback of requiring substantial computational resources. This is due to the need to train a model for each policy, which becomes increasingly demanding as the number of policies increases.
In this work, we present an effective meta-policy learning framework that is capable of generalizing diverse objectives derived from randomized reward functions.

\subsection{Robust Autonomous Driving}

Some prior work tries to improve the robustness of driving policy by applying adversarial attacks. 
\citet{chen2020lane} employs adversarial agents to evaluate the robustness of autonomous driving models. \citet{ding2021semantically} present a method that incorporates domain knowledge to generate adversarial scenarios, thereby improving the robustness of autonomous driving systems. 
\citet{sharif2021adversarial} first identify failure states in autonomous driving agents by training adversarial driving agents, then retrain their autonomous agents with the adversarial inputs. 
However, the generated behaviors of these adversarial agents are often unrealistic or even unreasonable in real-world settings.
\citet{kontes2020high} utilizes domain randomization for robust policy transfer from simulation to the real world.
In this work, we use a diverse set of learned social agents' behaviors by randomizing their reward functions to improve the robustness of the ego driving policy.

\section{Problem Formulation} \label{sec:problem_formulation}

\begin{figure*}
    \centering
    \includegraphics[width=0.9\textwidth]{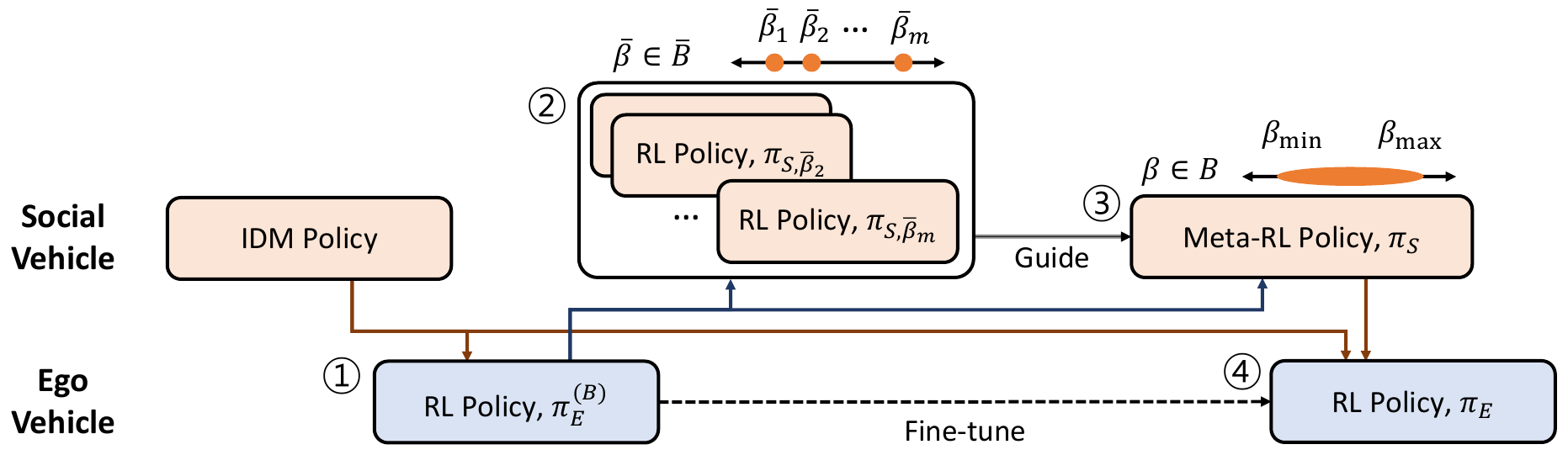}
    \vspace{-0.1cm}
    \caption{A diagram of the proposed training framework for social driving meta-policy and robust ego driving policy. The orange and blue boxes represent the social and ego agents' policies, respectively. The orange and blue lines indicate the setting of the training environment for a certain type of agent. (1) Specifically, the ego agent's policy is initially trained with IDM-based social vehicles as its training environment. (2) Then, the guiding policies with a limited preference set are trained, (3) followed by training a meta-policy on a broader preference set. (4) Finally, the ego agent's policy is trained in the environment that involves social vehicles with both the meta-RL policy and the IDM policy.}
    \vspace{-0.5cm}
    \label{fig:img1}
\end{figure*}

In an uncontrolled T-intersection scenario, as depicted in \cref{fig:intro},
multiple vehicles on a two-lane roadway are driving horizontally and one ego vehicle on a vertical road is trying to merge into the upper lane. In this situation, the primary objective for each vehicle is to reach its destination while minimizing the risk of collision. We formulate the uncontrolled T-intersection scenario as a Partially Observable Stochastic Game (POSG, \cite{posg}) representing the discrete-time stochastic control process. In this work, the POSG is defined as a $(\mathcal{I}, \mathcal {S}, \boldsymbol{\mathcal {O}}, \boldsymbol{\mathcal {A}}, \boldsymbol{\mathcal {R}}, T)$ tuple.

\subsubsection{Agent} $\mathcal{I}$ is the set of agent indices. We define an individual vehicle as a decision-making agent. Each agent is indexed with $i\in \mathcal{I} = \mathcal{I}_E \cup \mathcal{I}_S$, where $\mathcal{I}_E=\{0\}$ is the index set of ego vehicle that is trying to merge into the upper road and $\mathcal{I}_S=\{1, \dots, n\}$ is the index set of social vehicles that drive horizontally.

\subsubsection{State} $\mathcal{S}$ is the set of states. The physical state of each agent $x^i\in \mathbb{R}^4$ is composed of its position and speed. Additionally, the social agents have a preference level $\beta^i \in \mathbb{R}$ that regulates the degree of aggressiveness with respect to the ego agent. The global state $s\in\mathcal{S}$ is defined as
\begin{equation}
    s=\left[x^0,(x^1,\beta^1),(x^2,\beta^2),\dots,(x^n,\beta^n)\right].
\end{equation}

\subsubsection{Observation} $\boldsymbol{\mathcal{O}}=\mathcal{O}^0\times\mathcal{O}^1\times\cdots \times\mathcal{O}^n$ is the set of joint observations of all agents. In real-world scenarios, it is not realistic to directly observe the internal preference of surrounding vehicles. Therefore, the ego agent can only observe the physical state of each agent and infer their internal preference using an inference model. However, to simplify the environment, we assume that the social agents can access the true global state, which includes internal preference.
\begin{equation}
    o^i=
    \begin{cases}
        [x^0,x^1,\dots,x^n] & \text{if } i\in \mathcal{I}_E, \\[3pt]
        [x^0,(x^1,\beta^1),\dots,(x^n,\beta^n)] & \text{if } i\in \mathcal{I}_S.
    \end{cases}\label{eq:observation}
\end{equation}

\subsubsection{Action}
$\boldsymbol{\mathcal{A}}=\mathcal{A}^0\times\mathcal{A}^1\times\cdots \times\mathcal{A}^n$ is the set of joint action space of all agents. Action space, $\mathcal{A}^i$ is defined by a set of candidate speeds, represented by $\mathcal{A}^i=\{0.0, 0.5, 3.0\}\text{m/s}$ for $i\in\mathcal{I}$. The action of each agent controls the desired speed of its own low-level controller.

\subsubsection{Reward}
$\boldsymbol{\mathcal{R}}:\mathcal{S}\times\boldsymbol{\mathcal{A}}\to\mathbb{R}^{N}$ is the reward for each agent. The base reward $r^i$ for each vehicle is designed to encourage the learning agent to navigate the intersection with maximum speed and minimum collision risk, which is defined as
\begin{equation}
    r^i(s,\textbf{a})= 
    \begin{cases}
        r_{\text{goal}}          & \text{if } s\in S_{\text{goal}}^i,\\[3pt]
        r_{\text{fail}}          & \text{if } s\in S_{\text{fail}}^i,\\[3pt]
        r_{\text{speed}}\times \|v^i\|    & \text{otherwise},
    \end{cases}
\end{equation}    
where $S_{\text{goal}}^i$ is a state set indicating success cases where agent $i$ has reached its goal, while $S_{\text{fail}}^i$ is a state set indicating failure cases where a collision happens or the vehicle goes off the road. $v^i$ represents the speed of agent $i$.

The ego agent uses its base reward as a final reward denoted by $R^i$. However, social agents use a final reward which is defined as the sum of its own base reward and a base reward of ego weighted by its preference:
\begin{equation}\label{eq:final_reward}
    R^i(s,\textbf{a})=
    \begin{cases}
        r^0(s,\textbf{a}) & \text{if } i\in \mathcal{I}_E, \\[3pt]
        r^i(s,\textbf{a})+\beta^i r^0(s,\textbf{a}) & \text{if } i\in \mathcal{I}_S,
    \end{cases}
\end{equation}  
where $\beta^i$ denotes the preference of agent $i$.
We can manipulate the level of aggressiveness in the policy objective by modifying the preference using this reward design. For instance, a negative $\beta$ value will encourage minimizing the reward of the ego vehicle, preventing it from making a left turn. A positive $\beta$ value will encourage maximizing the reward of the ego vehicle, which encourages the social vehicles to yield.

\subsubsection{Transition}
$\mathcal{T}:\mathcal{S}\times\boldsymbol{\mathcal{A}}\to\mathcal{S}$ is a function that determines the next state given the current state and action. The transition model of the simulation operates with a time interval $\Delta t$ of 0.1s. Each social vehicle is assigned a sequence of straight waypoints that leads to the end of the road, while the ego vehicle has a sequence of straight waypoints initially followed by curved waypoints when merging into the upper road. The speeds of all vehicles are updated using low-level controllers and actions that follow the waypoints. The position of each vehicle is deterministically updated based on their previous positions $(p^x_{t}, p^y_{t})$ and speeds $(v^x_{t}, v^y_{t})$ as follows:
\begin{align}
    &p^x_{t+1}=p^x_{t}+v^x_t \cdot \Delta t ,\\
    &p^y_{t+1}=p^y_{t}+v^y_t \cdot \Delta t.
\end{align}

\section{Method}\label{sec:methods}

We aim to train an ego agent that can effectively interact with a diverse group of social agents and generalize to unseen situations.
The objective of the ego agent is formulated as
\begin{equation} \label{eq:ego_objective}
    \pi^*_{E} = \argmax_{\pi_E\in \Pi_E} \sum_{\pi_{S}\in\Pi_{S}} \mathbb{E}_{s_t, \textbf{a}_t}\left[
    \sum_{t=0}^{\infty} \gamma^{t}R(s_t,\textbf{a}_{t})\right],
\end{equation}
where $\pi_E$ and $\pi_S$ represent the policies for the ego and social agent, respectively. Similarly, $\Pi_E$ and $\Pi_S$ represent the feasible policy sets for the ego and social agent, respectively.
$\gamma$ represents the discount factor.
The initial state $s_0$ is sampled from the initial state distribution $\rho(\cdot)$. At each time step $t$, the action $\textbf{a}_t$ is sampled using the policies for both the ego and social agents, and the next state $s_t$ is sampled using the transition function based on the previous state and action.
The proposed method consists of two major steps. 
The first step is to learn diverse RL policies of social agents using the reward functions designed to emphasize interactions with the ego agent, which is achieved by our proposed meta-RL method effectively. 
The second step is to learn an ego agent's policy that identifies the internal preference of each social agent based on their past behavior and makes decisions accordingly. The overall training process is illustrated in \cref{fig:img1}.

\subsection{Training a Social Driving Meta-Policy}\label{subsec:meta}

While the reward design described in \cref{eq:final_reward} provides flexibility in defining various objectives, 
we found that training a policy directly on the complete continuous spectrum of preferences fails to yield rational aggressive behaviors.
To address the issue, we propose a two-stage meta-policy learning method, which enables the learned social agents' policy to generate diverse behaviors with a wide range of preferences. 

In the first stage, in order to train a guiding policy $\pi_{S,\beta}(a|o)$ that corresponds to a specific preference $\beta$, the objective of guiding policy for $\beta$ is written as
\begin{align}
        &\pi^{*}_{S,\beta} 
        = \argmax_{\pi_S\in \Pi_S} \mathbb{E}_{s_t, \textbf{a}_t}\left[\sum_{t=0}^T \gamma^{t}R_t^i(s_t,\textbf{a}_t) \right] \nonumber\\
        &\ \ \ =\argmax_{\pi_S\in \Pi_S} \mathbb{E}_{s_t, \textbf{a}_t}\left[\sum_{t=0}^T \gamma^{t}(r^i_t(s_t,\textbf{a}_t)+\beta r^0_t(s_t,\textbf{a}_t)) \right].
\end{align}
We train multiple guiding policies based on a limited set of preferences $\bar{B}=\{\bar{\beta}^1,\dots,\bar{\beta}^m\}$ using a model-free RL algorithm, PPO \cite{schulman2017proximal}, resulting in a total of $m$ guiding policies $\pi^{*}_{S,\bar{\beta}^1},\dots,\pi^{*}_{S,\bar{\beta}^m}$.

In the second stage, we train a meta-policy $\pi_{S}(a|o,\beta)$ that can generalize the behavior according to its preference. 
Unlike the first stage where the policy is trained on a limited set of preferences, the meta-policy is trained on a broader range of preference sets $B=\{\beta \ | \ \beta_{\text{min}}\leq\beta\leq\beta_{\text{max}}\}$. 
Learning a meta-policy that can simultaneously handle a wide range of preferences is challenging. 
To achieve this goal, we propose to apply regularization techniques to the meta-policy to mimic the behaviors of the guiding policies for the pre-trained preferences. 
This approach enables the meta-policy to learn behaviors with new preferences efficiently while retaining the ability to perform well with the preferences for guiding policies.
The regularization for guiding policies is
\begin{equation}\label{eq:regularization}
    \mathcal{L}_{\text{reg}}(\theta)=
    \sum_{\bar{\beta}\in \bar{B}}\mathds{1}\left(|\bar{\beta}-\beta|\leq d\right)D_{\text{KL}}\left(\pi_{S,\bar{\beta}}^*(\cdot|o) \| \pi_S(\cdot|o,\beta)\right),
\end{equation}
where $d$ denotes the guide distance.
When preferences are sampled from a continuous space, as opposed to a discrete space in the first stage, it becomes infeasible to sample preferences within a limited preference set $\bar{B}$. Therefore, if the sampled preference $\beta$ is sufficiently close to any preference $\bar{\beta}\in \bar{B}$, we encourage the meta-policy $\pi_S$ to mimic the guiding policy $\pi_{S,\bar{\beta}}^*$ as a regularization strategy. 

Finally, the parameter of meta-policy is updated using a weighted sum of the PPO loss and the regularization loss in \cref{eq:regularization}, which is written as
\begin{equation}
    \mathcal{L}(\theta)= \mathcal{L}_{\text{PPO}}(\theta)+w_\text{reg}\mathcal{L}_{\text{reg}}(\theta),
\end{equation}
where $w_\text{reg}$ denotes the weight for the regularization loss.

To facilitate the learning of the social policy, a rational ego vehicle's behavior is necessary. 
Since designing the ego behavior based on pre-defined rules can be challenging and may not generalize well, we adopt an RL-based ego driving policy, $\pi_E^{(B)}$, obtained by the method in \cite{liu2022learning}. This RL-based policy has demonstrated effective interactions with social vehicles controlled by the Intelligent Driver Model (IDM) \cite{treiber2000congested}.

\subsection{Training a Robust Ego Driving Policy}

In this section, we present our approach to enhance the robustness of the ego agent's policy to interact with various types of social vehicles.
Existing methods \cite{ma2020latent, liu2022learning} employed an IDM-based policy for social vehicles in the training environment and evaluated the performance using the same one. 
Although these methods have shown successful interactions with IDM-based social vehicles, we have observed limitations in the ability of the ego agent's policy to interact with unseen social behavior policies.
To address this issue, we incorporate both the social vehicles used in existing methods and the RL-based social vehicles learned in \cref{subsec:meta} in the training environment for ego policy learning.
This mixed environment aims to enhance the adaptability and performance of the ego agent in scenarios with diverse social driving behaviors.

The ego agent should navigate the intersection efficiently and minimize the risk of collision with the social vehicles. 
To achieve this goal, it is necessary to understand the internal preferences of the social vehicles, which indicate to what extent they are willing to yield to the ego agent. The internal preferences can be inferred by analyzing the past trajectories of the vehicles.
However, relying solely on the reward signal to extract the internal preferences requires a substantial number of training samples. To address the issue, we adopt an unsupervised learning approach for social vehicles introduced in \cite{liu2022learning}, which employs a GNN-GRU structure that can effectively process spatio-temporal data. The internal preference of each social vehicle can be appropriately extracted by an auxiliary task which is to reconstruct the historical trajectories.

\begin{figure*}
    \centering
    \includegraphics[width=\textwidth]{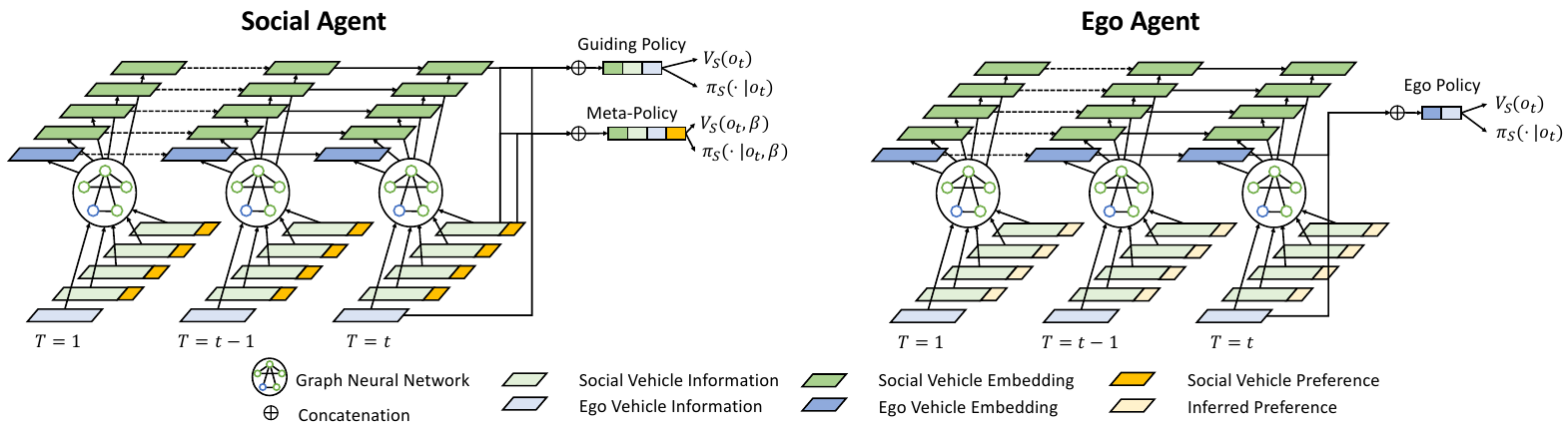}
    \vspace{-0.5cm}
    \caption{
    The policy and value network architectures for the social agents (left) and the ego agent (right).
    }
    \vspace{-0.4cm}
    \label{fig:architecture}
\end{figure*}

The ego agent enhances its policy by concatenating the latent features obtained from the inference network with the observations of each vehicle. Unlike the social agent, which exhibits diverse objectives ranging from aggressiveness to conservatism, the ego agent has a single objective that does not necessitate a meta-policy. The value and policy functions of the ego agent are updated using the PPO algorithm, similar to the social agents.
Training an inference network together with the policy and value networks can lead to unstable training, as the output of the inference network is fed into the input of the other networks. 
Therefore, we pre-train the inference network and subsequently train the policy and value networks using the pre-trained inference network, following a widely used approach in representation learning for RL.

\subsection{Network Architecture for Ego and Social Agents}

The architecture for the policy and value network of social agents is described in \cref{fig:architecture}. 
We construct the observation of each agent as a graph where the nodes represent the surrounding vehicles, including the agent itself, and the edges are formed by connections originating from all other vehicles to the agent. 
For the ego agent, the node feature of surrounding vehicles is formed by concatenating the physical state and the latent feature derived from the historical trajectory. 
For social agents, the node feature of surrounding vehicles is formed by concatenating the physical state and the preference as described in \cref{eq:observation}.
The preference of the ego agent is padded with zeros to maintain the same size. Next, we use a GNN to extract essential features from the constructed graph. These extracted features are then fed into a GRU network. Subsequently, the hidden features obtained through this process are passed through two MLP networks to obtain policy and value, respectively. The hidden feature of the social meta-policy includes an extra concatenation of its own preference value, while the ego policy solely concatenates its own embedding and information.
We maintain the same structure for the ego agent as described by \citet{liu2022learning} to evaluate the effectiveness of our training framework. 
Since we need to train several guiding policies simultaneously, we adopt a strategy to share the GNN and GRU networks among the guiding policies and separate the final MLP networks for higher training efficiency.

\section{Experiments}
    \label{sec:experiments}

\begin{figure}
    \centering
    \includegraphics[width=\columnwidth,height=3.5cm]
    {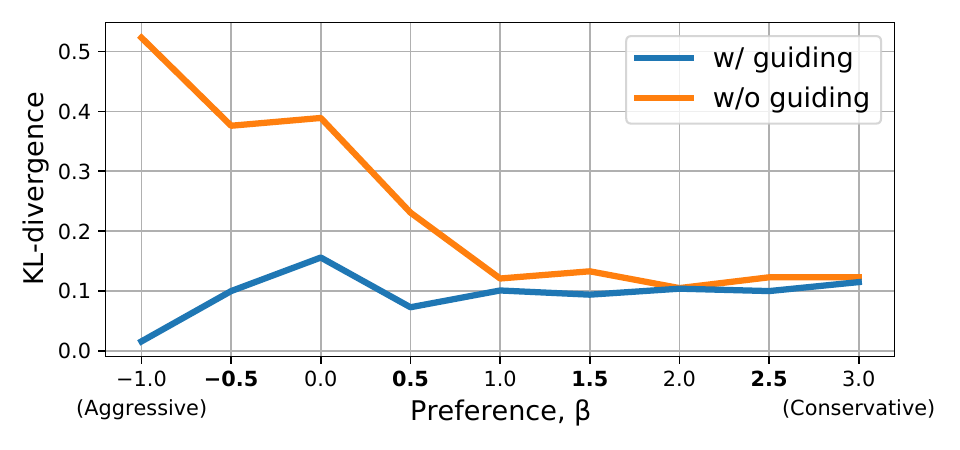}
    \vspace{-0.7cm}
    \caption{KL-divergence between the meta-policy and guiding policies for social vehicles. 
    The comparison is conducted with an interval of 0.5.}
    \vspace{-0.53cm}
    \label{fig:exp1-kldiv}
\end{figure}

\subsection{Simulation and Training Settings}
We train the social policies in two stages and train the ego agent's policy using 10M samples. 
The learning rate for the PPO algorithm is set to \num{e-4} with a linear decay. In the first stage, the guiding policies are trained with a preference set of $\bar{B}=\{-1,0,\dots,+3\}$. In the second stage, the meta-policy uses a preference as a real value ranging from \num{-1} to \num{3}. 
When the social vehicle is generated, the preference value is uniformly sampled from the preference set in both stages.
We set the guide distance $d$ of the social meta-policy to \num{0.1} and the weight for the regularization loss $w_{\text{reg}}$ to \num{0.01}. 
Consequently, there was a \num{20}\% probability that a sample from the social policy was guided.
To train the ego policy, we randomly select IDM or RL social policies to control social vehicles with an equal probability to generate training scenarios.

\subsection{Validation of Social Driving Meta-Policy}\label{sec:expb}

\begin{figure}
    \centering
    \includegraphics[width=\columnwidth,height=3.5cm]{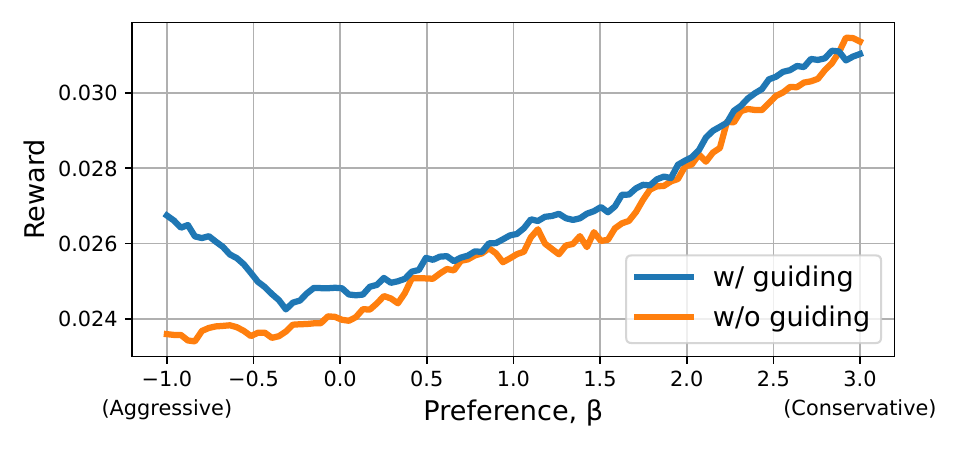}
    \vspace{-0.73cm}
    \caption{The reward of social meta-policy according to the preference. The blue and red line represents the proposed and ablation method, respectively. Note the significant improvement by guiding policies for aggressive agents.}
    \vspace{-0.4cm}
    \label{fig:exp1}
\end{figure}

\begin{figure}
    \centering
    \includegraphics[width=\columnwidth,height=3.5cm]{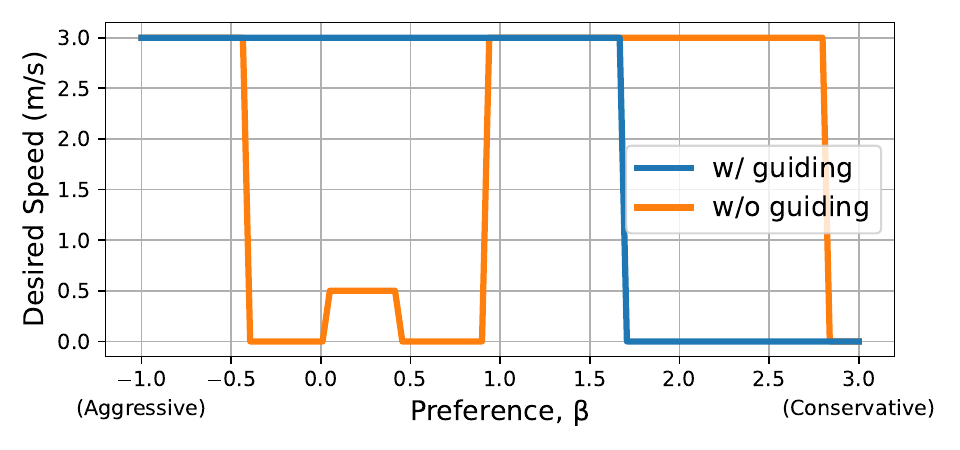}
    \vspace{-0.73cm}
    \caption{
    The desired speeds (i.e., action) determined by the social meta-policy as preference value changes in the situation are illustrated in \cref{fig:intro}.}
    \vspace{-0.5cm}
    \label{fig:scenario}
\end{figure}

\begin{figure*}[!tbp]
    \centering
    \includegraphics[width=\textwidth]{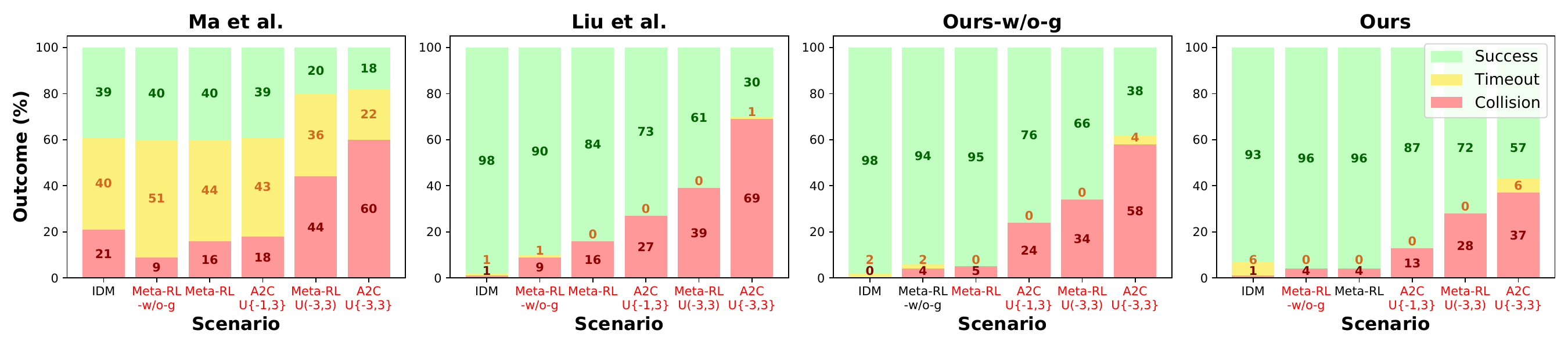}
    \vspace{-0.7cm}
    \caption{The cross-evaluation results among different types of ego and social agents. 
    The red text represents the out-of-distribution scenarios. 
    }
    \vspace{-0.4cm}
    \label{fig:cross_eval}
\end{figure*}

\begin{figure*}[!tbp]
    \centering
    \includegraphics[width=\textwidth]{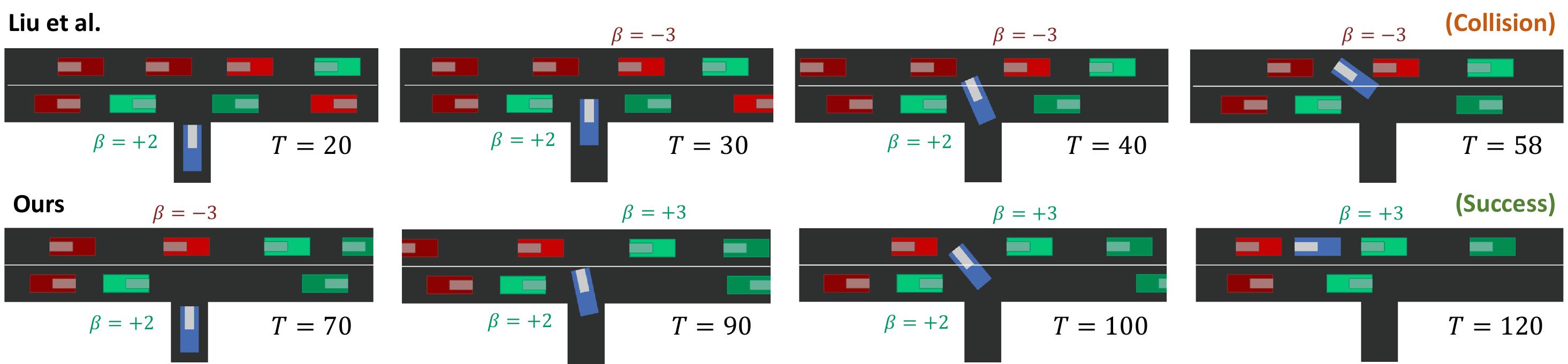}
    \vspace{-0.5cm}
    \caption{Qualitative results in an OOD A2C-U\{-3,3\} challenging scenario. Both methods are tested in exactly the same scenario. The ego car is represented in blue, aggressive social cars in red and conservative social cars in green. The darkness of colors indicates the degree of conservativeness or aggressiveness. \citet{liu2022learning}'s ego agent can be observed attempting to proceed through the intersection aggressively, as depicted in the second subfigure ($T=30$). Despite the narrow gap between the social vehicles, it still tries to proceed, resulting in a collision. On the other hand, our ego agent waits for the aggressive vehicles in the above lane to pass the intersection before taking action. Once both incoming vehicles are conservative, it finally attempts to proceed through the intersection, which allows for safer navigation and avoids collision. Best viewed in color.}
    \vspace{-0.4cm}
    \label{fig:visualization}
\end{figure*}

In this subsection, we evaluate the ability of our proposed meta-RL method to generate behaviors according to diverse preferences. We compare our proposed method with the non-regularized meta-policy.
First, we assess the similarity between the meta-policy and the guiding policies under various preferences to verify whether the meta-policy exhibits rational behavior even without explicit guidance. 
Specifically, we estimate the KL divergence of the policy distribution for each preference $D_{\text{KL}}(\pi_{S,\beta}^*(\cdot\mid o)||\pi_S(\cdot\mid o,\beta))$ with 100K samples. 
\cref{fig:exp1-kldiv} illustrates the KL divergence between the meta-policy and guiding policy for various preferences. The bold values indicate preferences that were not employed during the training of the meta-policy.
The results indicate that using guiding policies leads the meta-policy to align its policy distribution with guiding policies.
However, without the guided training strategy, the meta-policy exhibits an increasing deviation from the guiding policies as the preference value decreases, which implies that the meta-policy without guided training fails to generate the desired aggressive behaviors.

\cref{fig:exp1} illustrates the performance of the meta-policy across a wide range of preference settings, showing their ability to improve the rewards. We collect 3M samples for each method to evaluate the performance. It demonstrates that our proposed method with guiding policies achieves better performance than the ablation method, especially in cases with aggressive preferences.
The difference between the meta-policies trained with and without guiding policies is particularly significant during interactions with the ego agent. When the ego agent attempts to merge into the upper lane, the behavior of the social agent exhibits variability due to different preferences. \cref{fig:scenario} shows the actions taken by the meta-policy corresponding to different preferences. The meta-policy learned with guiding policies exhibits a rational behavior by prioritizing passing when the preference value is low and yielding when the preference value is high. However, the meta-policy without guiding generates behaviors that are inconsistent with preferences.

\subsection{Cross-Evaluation of Ego Driving Policy}

We evaluate the robustness of the ego policy based on its interactions with various types of social vehicles. We design the test scenarios to focus on challenging situations which involve diverse driving styles, presenting realistic driving scenarios for the ego agent. In different testing scenarios, the social vehicles are controlled by the following policies.
\begin{itemize}[leftmargin=*]
    \item \textbf{IDM}: An IDM-based social policy used in baseline methods, characterized by aggressive and conservative behaviors.
    \item \textbf{Meta-RL}: A social policy designed to generalize across various preferences, trained using the proposed method.
    \item \textbf{Meta-RL-w/o-g}: A social meta-policy designed to generalize across various preferences, trained without guiding policies. It is an ablation method of Meta-RL.
    \item \textbf{A2C-U\{-1,3\}}: A set of social policies tailored for specific preferences, covering a range of preferences using a different RL algorithm A2C \cite{mnih2016asynchronous}. The preference values are sampled from the set $\{-1, +0, \dots, +3\}$.
    \item \textbf{Meta-RL-U(-3,3)}: A meta-RL social policy where the preference of each social vehicle is sampled from -3 to 3. It generates more challenging scenarios than Meta-RL.
    \item \textbf{A2C-U\{-3,3\}}: An A2C-based social policy where the preference is uniformly sampled from $\{-3, -2, \dots, +3\}$. It generates more challenging scenarios than A2C-U\{-1,3\}.
\end{itemize}

We compare the performance of the ego agent's policy trained with the following methods.
\begin{itemize}[leftmargin=*]
    \item \textbf{\citet{ma2020latent}}: An ego policy trained with IDM social policy. 
    \item \textbf{\citet{liu2022learning}}: An ego policy trained with IDM social policy.
    \item \textbf{Ours-w/o-g}: An ego policy trained with IDM and Meta-RL-w/o-g social policies.
    \item \textbf{Ours}: An ego policy trained with IDM and Meta-RL social policies.
\end{itemize}

\cref{fig:cross_eval} shows the decision making performance of the ego agent's policy in various testing environments.
The bar charts show the collision, timeout, and success rates. Scenarios with red texts indicate out-of-distribution (OOD) scenarios that are not included during the training process of the corresponding ego policy.
The method of \citet{ma2020latent} shows significantly higher collision and timeout rates among baseline methods. 
The reason is that their method classifies social vehicles into only two classes (i.e., conservative/aggressive), which cannot well capture the diversity of social behaviors.
The method of \citet{liu2022learning} achieves better performance when interacting with IDM-based social vehicles due to the flexibility in encoding social behaviors. However, it exhibits a high collision rate with RL-based social vehicles, which indicates that training only with the IDM social policy can result in overfitting certain social behaviors.
The ego policy of our ablation method (Ours-w/o-g) exhibits strong compatibility with the social vehicles encountered during training but demonstrates poor performance in OOD scenarios.
The ego policy of our method (Ours) interacts well with both the IDM and meta-RL social vehicles used for training. Moreover, the ego policy demonstrates better performance in interacting with social vehicles with OOD policies, which implies that the guided social meta-policy enhances the robustness of the ego policy.
\cref{fig:visualization} provides the visualization of a challenging OOD testing scenario where our method makes a successful left turn in the T-intersection while the baseline method causes a collision.

\section{Conclusions and Discussions}
    \label{sec:conclusion}
In this paper, we present a novel training strategy to enhance the robustness of RL-based autonomous driving by generating diverse social behaviors in training environments. 
We propose a guided meta reinforcement learning method to enable the generation of a variety of reasonable behaviors for social agents effectively. 
The meta-RL policy achieves effective action generation over a wide range of internal preferences that indicate the degree of aggressiveness. 
The ego agent's policy trained with the social meta-policy exhibits a remarkable level of robustness to OOD scenarios where social vehicles have unseen behavior styles.
In future work, we aim to ensure safety by incorporating safe RL and generalizing the traffic scenarios by leveraging real-world datasets.

\printbibliography

\end{document}